\documentclass[conference]{IEEEtran}
\IEEEoverridecommandlockouts
\usepackage{cite}
\usepackage{amsmath,amssymb,amsfonts}
\usepackage{algorithmic}
\usepackage{graphicx}
\usepackage{subcaption}
\usepackage{textcomp}
\usepackage{xcolor}
\usepackage{tabularray}
\usepackage{stackengine}

\newcommand{\algon}[1]{{\small {#1}}} 

\def\BibTeX{{\rm B\kern-.05em{\sc i\kern-.025em b}\kern-.08em
    T\kern-.1667em\lower.7ex\hbox{E}\kern-.125emX}}
\begin{document}

\title{\huge A Study on the Inductance and Thermal Regression and Optimization for Automatic Layout Design of Power Modules}


\author{\IEEEauthorblockN{Victor Parque, Aiki Nakamura, Tomoyuki Miyashita}
\IEEEauthorblockA{\textit{Department of Modern Mechanical Engineering} \\
\textit{Waseda University}\\
3-4-1 Okubo, Shinjuku, Tokyo, Japan \\
parque@aoni.waseda.jp, anakamura1999@fuji.waseda.jp, tomo.miyashita@waseda.jp}
}

\maketitle

\begin{abstract}
Power modules with excellent inductance and temperature metrics are significant to meet the rising sophistication of energy demand in new technologies. In this paper, we use a surrogate-based approach to render optimal layouts of power modules with feasible and attractive inductance-temperature ratios at low computational budget. In particular, we use the class of feedforward networks to estimate the surrogate relationships between power module layout-design variables and inductance-temperature factors rendered from simulations; and Differential Evolution algorithms to optimize and locate feasible layout configurations of power module substrates minimizing inductance and temperature ratios. Our findings suggest the desirable classes of feedforward networks and gradient-free optimization algorithms being able to estimate and optimize power module layouts efficiently and effectively.
\end{abstract}

\begin{IEEEkeywords}
power modules, neural networks, automatic layout design, system identification, optimization
\end{IEEEkeywords}

\section{Introduction}

Power modules are essential to meet the rising sophistication and demands for efficient and cost-effective conversion of electrical energy in new consumer and industrial technology. New materials, such as Silicon Carbide (SiC), and new circuit manufacturing methods, such as 3-dimensional wiring, have enabled new frontiers towards the smaller size and the higher speeds of switching while ensuring reasonable thermal loads. 

When designing the layout and the packaging structures of power modules, minimizing the (parasitic) inductance and the heat dissipation are relevant to meet loads reliably and safely\cite{mantooth18,nan21}. As such, approaches derived from gradient-free optimization techniques have attracted the attention of the community. For instance, \cite{bello21} use Genetic Algorithm (GA) and Dijkstra for layout optimization of a buck converter board. \cite{mei20} used GA to design layouts of silicon carbide (SiC) modules with low parasitic inductance and resistance and small footprint. \cite{ben17} used sequential quadratic programming (SQP) to optimize the number of LEDs and their placement in a substrate of a multi-chip LED package. Hao et, al. \cite{hao17} used NSGA-II with cycle-route limits and ANSYS Q3D for minimal parasitic inductance, Abe et. al.\cite{icep21} proposed a routing-based approach for three-dimensional power modules with minimal inductance, and Nakamura et. al. \cite{icep22} used NSGA-II considering inductance and thermal performance as goals. 

It is well understood that both (expert-based) iterative design and optimization frameworks are time and resource consuming. Also, when new layouts are to be designed, knowledge of previous iterations is lost in population-based schemes. In this paper, we aim at finding optimal configurations of design-layout variables of power modules by a surrogate-based approach through nonlinear regression and optimization machinery. Thus, our key contributions are:


\begin{figure}[t!]
\centering
\includegraphics[width=0.65\columnwidth]{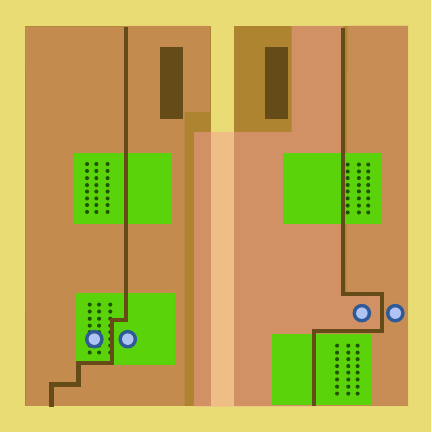}
\caption{Overview of the layout of a power module.}
\label{layout}
\end{figure}

\begin{itemize}
  
  \item We performed computational experiments outlining the regression capability of $n = 36$ parameters involving the layout-design of components of a class of power module (Fig. \ref{layout})\cite{icep22} to performance factors in terms of control path inductance ($f_1$), main path inductance ($f_2$), and maximum temperature in the substrate ($f_3$). Our results show the attractive regression performance of a class of network with 10 layers and 7580 nodes, achieving Mean Squared Error (MSE) in the order of $10^{-6}$ for $f_1$, $10^{-7}$ for $f_2$, and $0.19$ for $f_3$.
      
  \item We used the best obtained surrogate functions (neural networks) to optimize/locate the feasible/attractive layout configurations of power module substrates aiming to minimize a weighted-composition function of control path inductance, main path inductance, and maximum temperature in the substrate by using Differential Evolution machinery. Our computational experiments show the feasibility of convergence at feasible regions at low computational budgets.
  
\end{itemize}

\newpage

\section{Approach}


Our approach is inspired by the use of surrogates in expensive optimization problems\cite{neuro17}. We first aim at computing 

\begin{equation}\label{map}
\lambda_i: x \rightarrow f_i,
\end{equation}
in which $x \in \mathbb{R}^n$ is the set of observations of design variables of power module layouts, and $f_i$ is the set of observations of performance metrics of the power modules obtained from NSGA-II and design/simulation experiments\cite{icep22}. We consider the power module substrate studied in \cite{icep21} and \cite{icep22}, which considers the $n = 36$ parameters to determine the locations/layout of components and to minimize the control path inductance ($f_1$), the main path inductance ($f_2$), and the maximum temperature in the substrate ($f_3$). The mapping $\lambda_i$ can be computed by Feedforward Networks or Symbolic Regression schemes; and once reasonable mapping functions $\lambda_i$ are found, we aim at solving the following

\begin{equation}\label{eopt}
\begin{aligned}
& \underset{x}{\text{Minimize}}
& & F(x) = \sum_{i = 1}^{3} w_i\lambda_i(x) \\
& \text{subject to}
& & x \in \mathbf{X}
\end{aligned}
\end{equation} 
where $x \in \mathbb{R}^n$ is the optimization variable encoding the layout of the power module, $\mathbf{X}$ is the search space of feasible power modules, and $w_i$ are weights encoding the user-defined preference on the performance metric (inductances and temperature). Although the above-mentioned approach has been used in expensive optimization problems\cite{neuro17}, the use of surrogates in power module optimization has received little attention.

\section{Computational Simulations}

\begin{figure*}[t!]
\centering
\begin{tblr}{
    colspec = {|Q[c]|},
    rowsep = 3pt,
    }
    \hline[0.25em,gray3]
    \textbf{(a)} Learning performance. Dark (bright) colors imply neural networks with small (large) number of layers (refer to Table I).\\
    \hfill
    \stackunder{\includegraphics[width=0.32\textwidth]{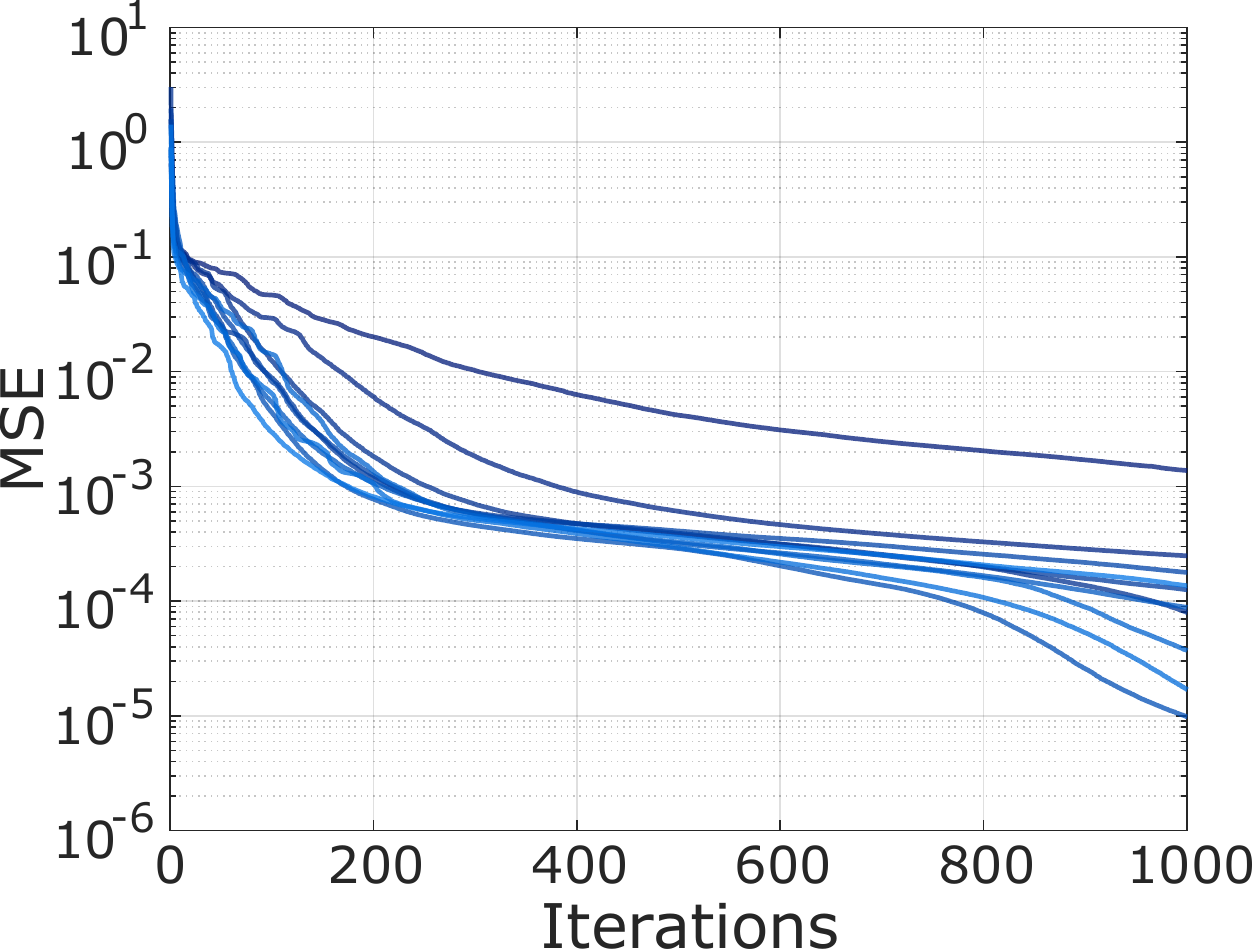}}{$f_1$}\hfill
    \stackunder{\includegraphics[width=0.32\textwidth]{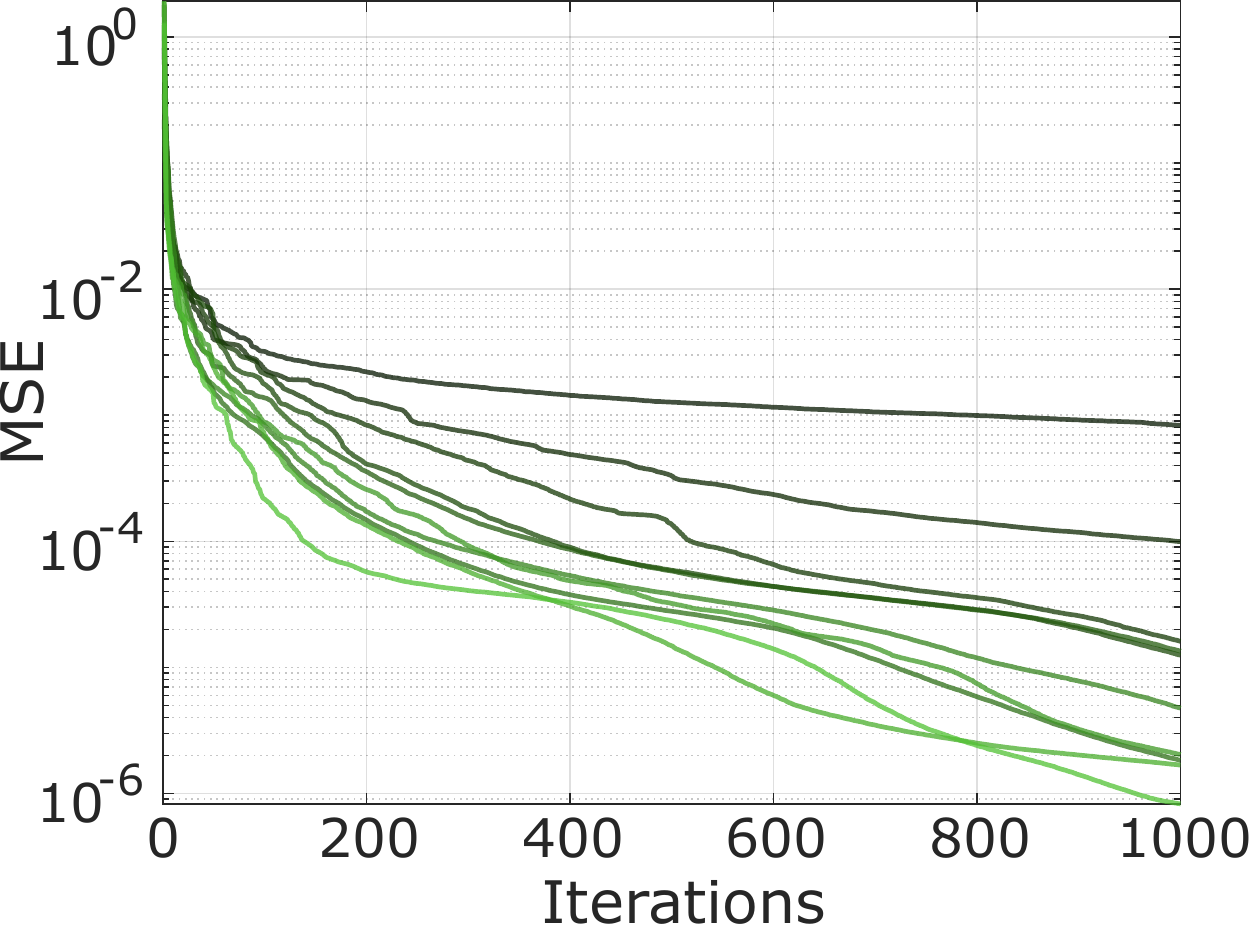}}{$f_2$}\hfill
    \stackunder{\includegraphics[width=0.32\textwidth]{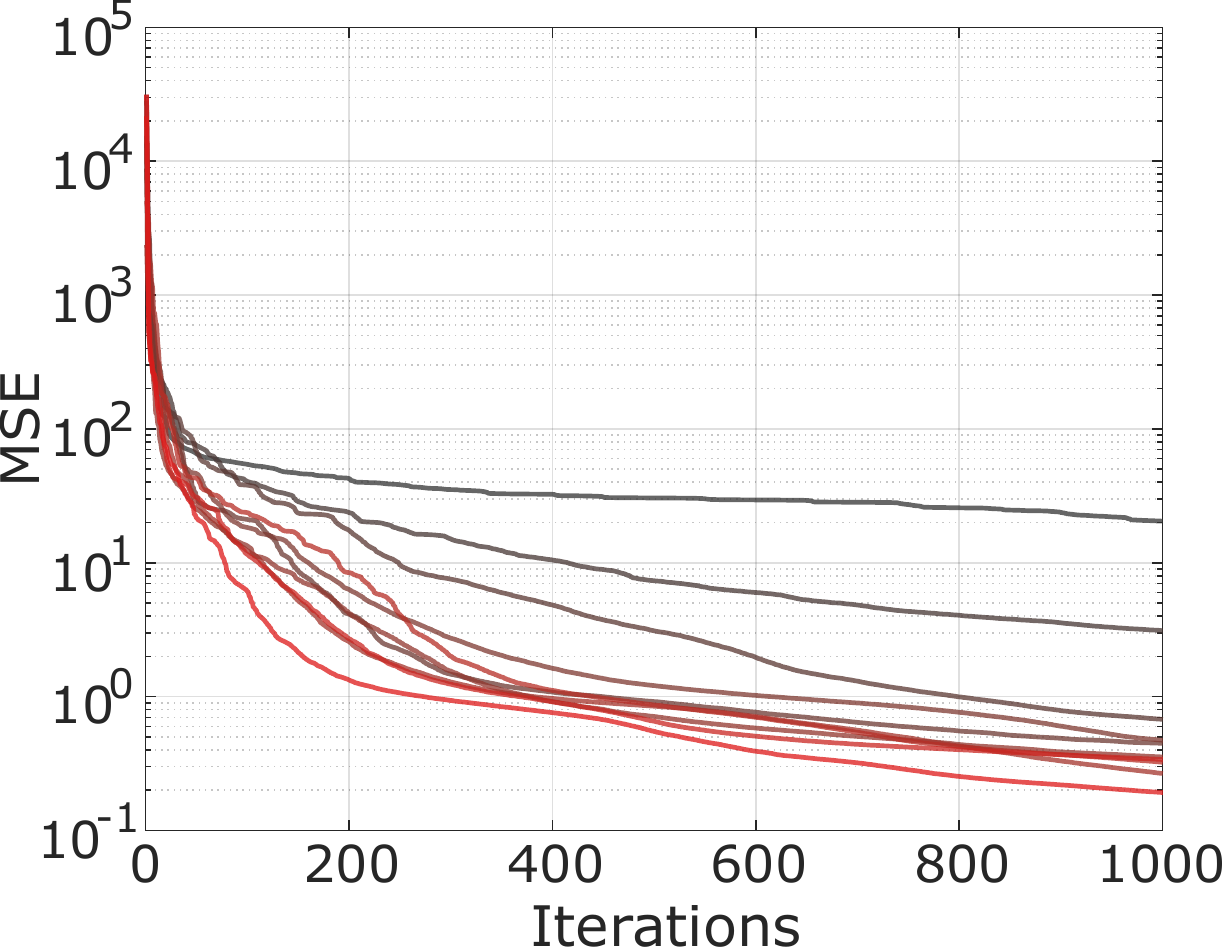}}{$f_3$}\hfill
    \hfill
	\\
    \hline[dotted]
    \textbf{(b)} Overview of training times. The numbers in the x-axis refer to neural networks from Table I.\\
    \hfill
    \stackunder{\includegraphics[width=0.32\textwidth]{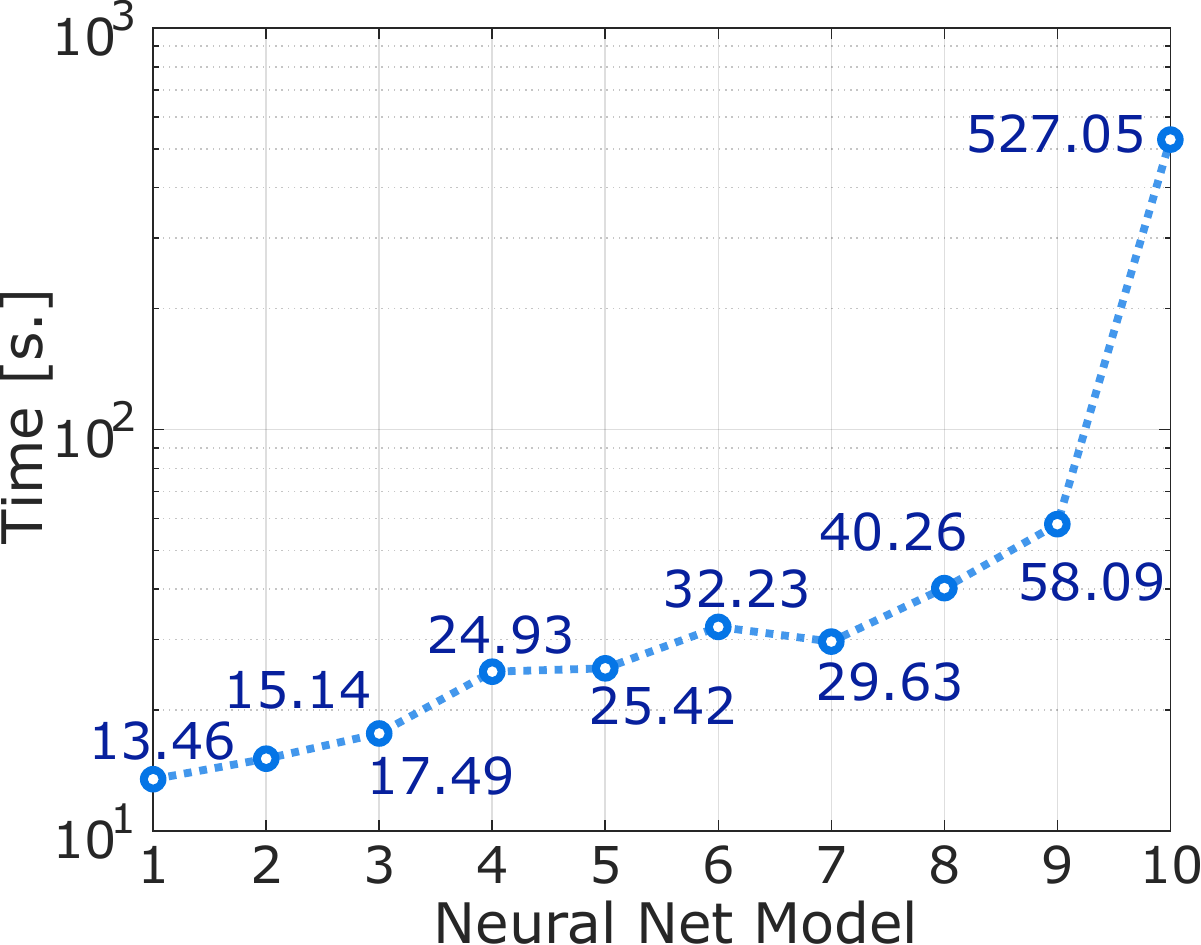}}{$f_1$}\hfill
    \stackunder{\includegraphics[width=0.32\textwidth]{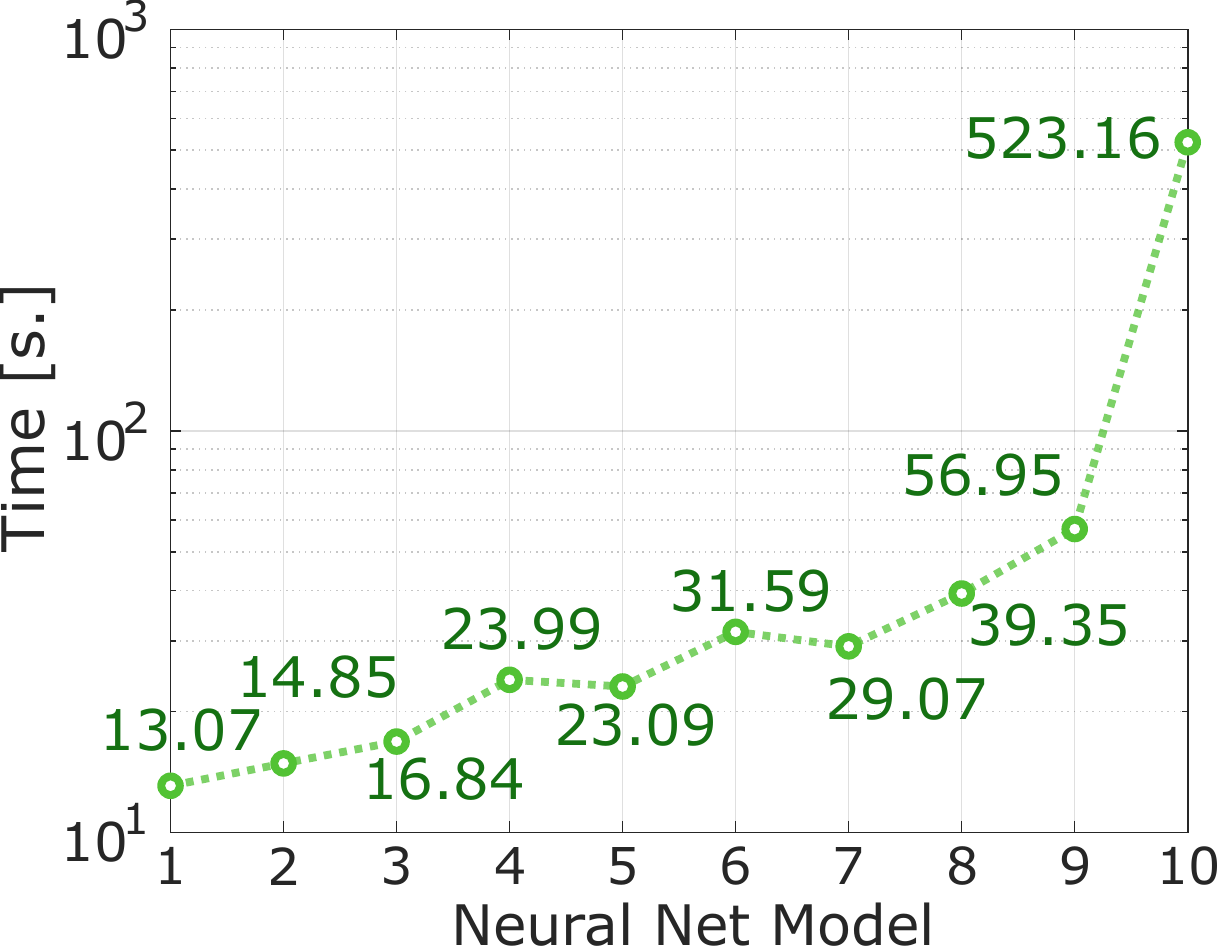}}{$f_2$}\hfill
    \stackunder{\includegraphics[width=0.32\textwidth]{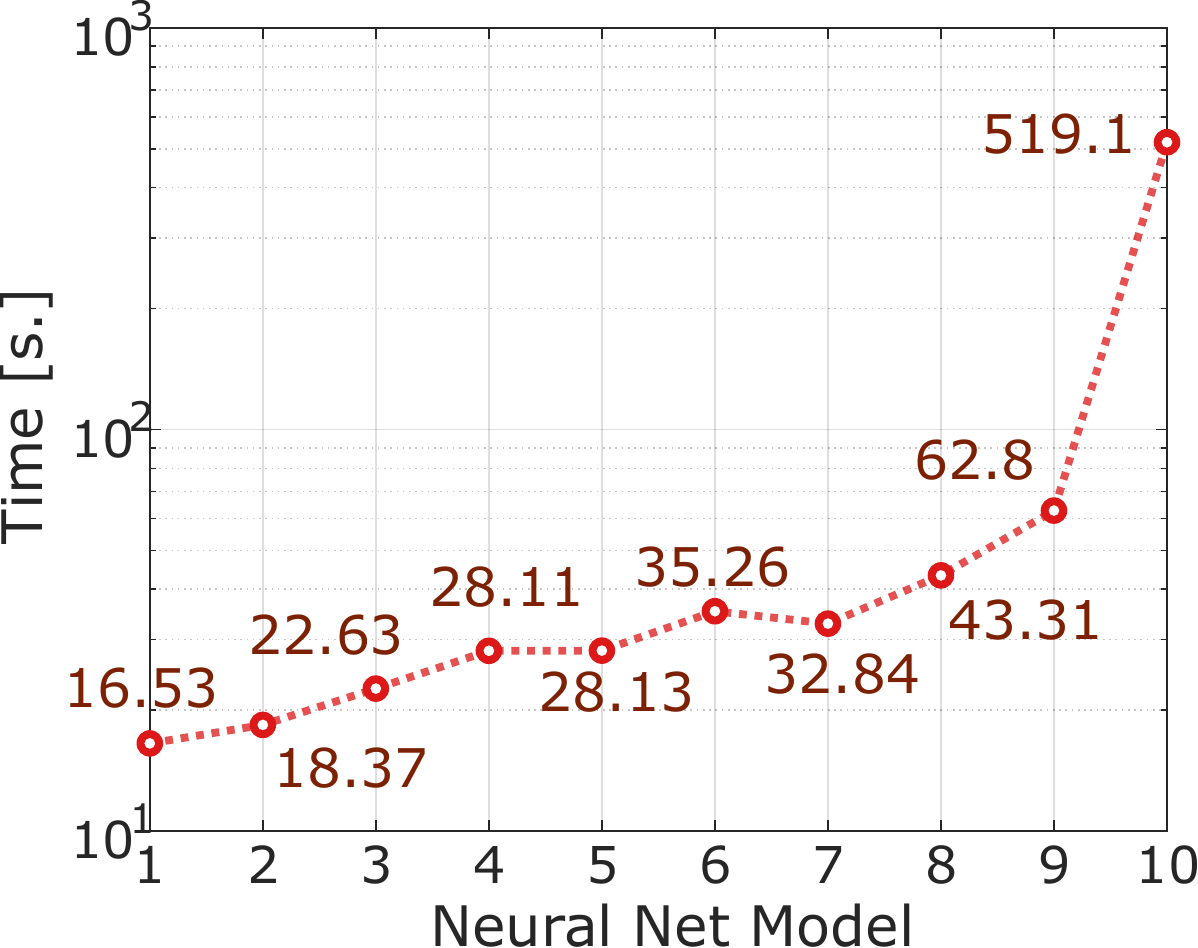}}{$f_3$}\hfill
	\\
    \hline[dotted]
    \textbf{(c)} Overview of prediction performance of best obtained models.\\
    \includegraphics[width=0.98\textwidth]{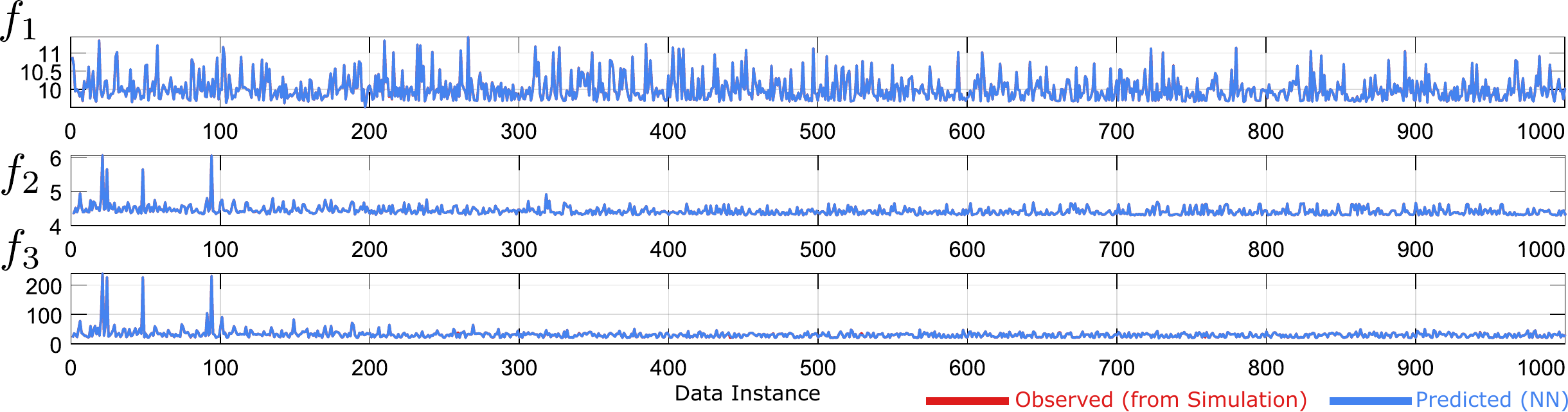} 
    \\
    \hline[0.25em,gray3]
\end{tblr}
\caption{Overview of (a) the learning convergence, (b) the training times, (c) the predictive ability of best obtained models. }
\label{resnn}
\end{figure*}

\begin{figure}[t!]
\centering
\includegraphics[width=0.98\columnwidth]{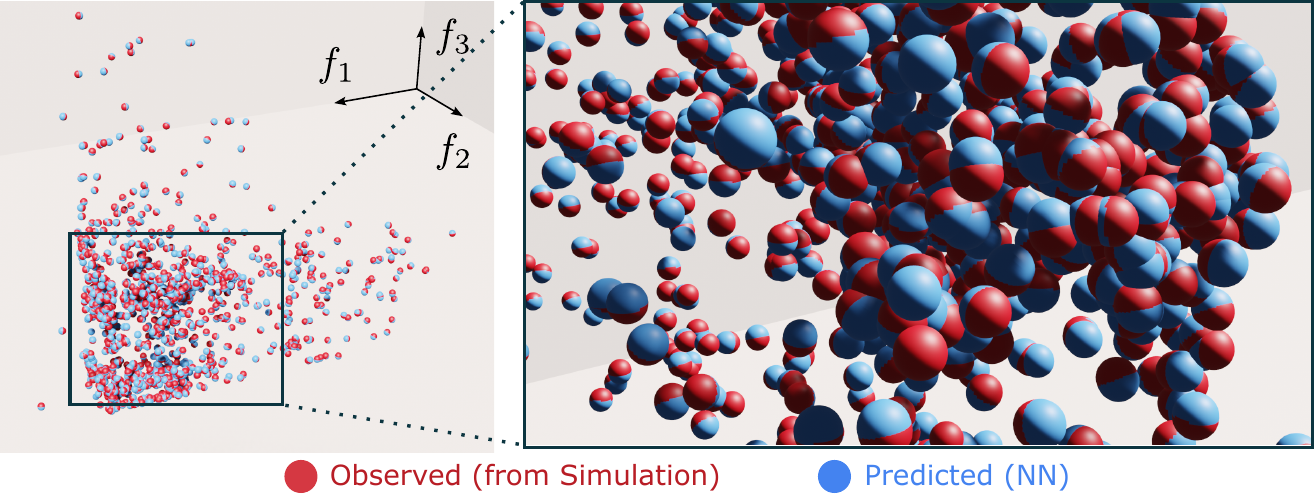}
\caption{Overview of performance metrics for Power Modules.}
\label{comp3d}
\end{figure}

In order to evaluate the feasibility of our approach, we tackle (\ref{map}) by a nonlinear regression scheme through Feedforward Networks with up to 7580 nodes distributed in up to 10 layers, as shown by Table \ref{nn} (best in bold). Although learning each of the neural network takes a few seconds/minutes, the convergence is competitive in the range of $10^{-3}$ to $10^{-1}$ MSE, as shown by Fig. \ref{resnn}, implying the reasonable approximation of control path inductance ($f_1$ at $9.78 \times 10^{-6}$), the main path inductance ($f_2$ at $8.25 \times 10^{-7}$), and maximum temperature in the substrate ($f_3$ at $0.19$) as shown by Fig. \ref{comp3d}.

As such, we used the best obtained performance functions $\lambda_i$ ($i \in [1,3]$) in terms of neural network schemes to tackle (\ref{eopt}) by population-based gradient free optimization algorithms. For rigorous evaluations, we use the class of algorithms inspired by Differential Evolution considering diverse forms of selection, mutation and adaptation mechanisms:

\begin{itemize}
  \item \small \algon{DERAND}: DE/rand/1/bin Strategy\cite{de97},
  \item \small \algon{DEBEST}: DE/best/1/bin Strategy\cite{de97},
  \item \small \algon{DESPS}: DE with Speciation Strategy\cite{desps},
  \item \small \algon{SHADE}: Success-History based Adaptive DE\cite{shade},
  \item \small \algon{RBDE}: Rank-based Differential Evolution\cite{rbde},
  \item \small \algon{JADE}: Adaptive DE with External Archive\cite{jade},
  \item \small \algon{DEGL}: DE with Local and Global Neighborhoods\cite{degl},
  \item \small \algon{DESIM}: DE with Similarity Based Mutation\cite{desim},
  \item \small \algon{DCMAEA}: Differential CMAE\cite{dcmaea},
  \item \small \algon{OBDE}: Opposition Based DE\cite{obde},
\end{itemize}

\begin{table}[t]
\centering
\caption{Configuration of Neural Net models.}\label{nn}
\NewTableCommand\myhline{\hline[0.1em,red5]}
\scalebox{0.98}{
\begin{tblr}
[
caption = {List of neural net models},
entry = {Short Caption},
label = {tbnn},
]
{
colspec = {cl},
}
\myhline
Neural Model & Number of nodes, and layer composition \\
\myhline
\textsf{1} &20 10  \\ 
\textsf{2} &50 20 10 \\ 
\textsf{3} &100 50 20 10 \\  
\textsf{4} &500 100 50 20 10 \\  
\textsf{5} &500 100 20 10  \\
\textsf{6} &1000 100 50 20 10 \\  
\textsf{7} &1000 100 10  \\
\textsf{8} &400 300 200 100 50 20 10 \\  
\textsf{9} &1000 300 200 100 50 20 10  \\
\textsf{\textbf{10}} &\textbf{5000 1000 500 400 300 200 100 50 20 10} \\
\myhline
\end{tblr}}
\vspace{-0.5cm}
\end{table}

\begin{figure*}[t!]
\centering
\begin{tblr}{
    colspec = {|Q[c]|},
    rowsep = 3.8pt,
    }
    \hline[0.25em,gray3]
    \textbf{(a)} Scenario 1: $w_1 = 1, ~ w_2 = 1, ~ w_3 = 2$.\\
    \hfill
    \stackunder{\includegraphics[width=0.32\textwidth]{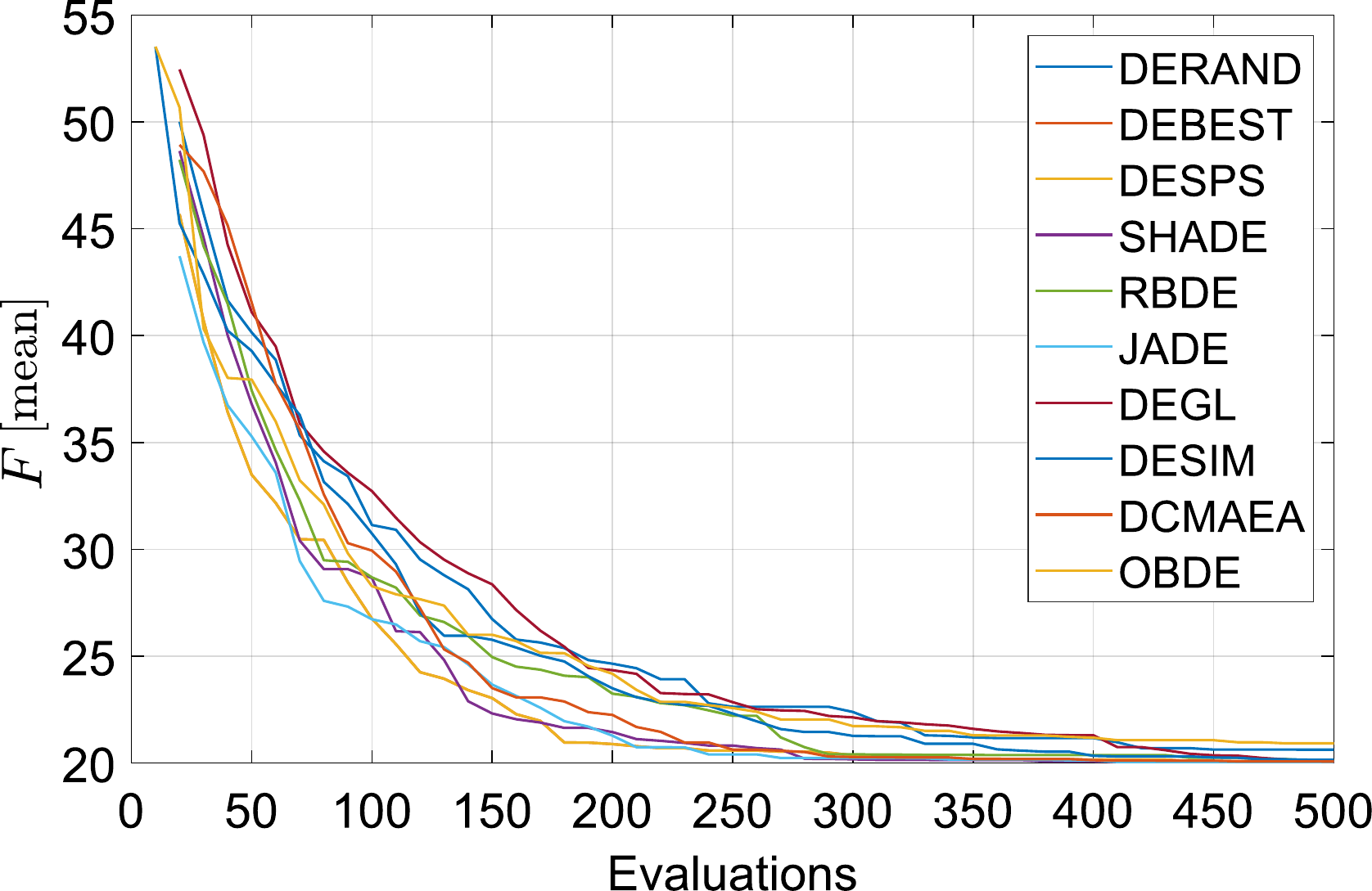}}{Mean Convergence}\hfill
    \stackunder{\includegraphics[width=0.32\textwidth]{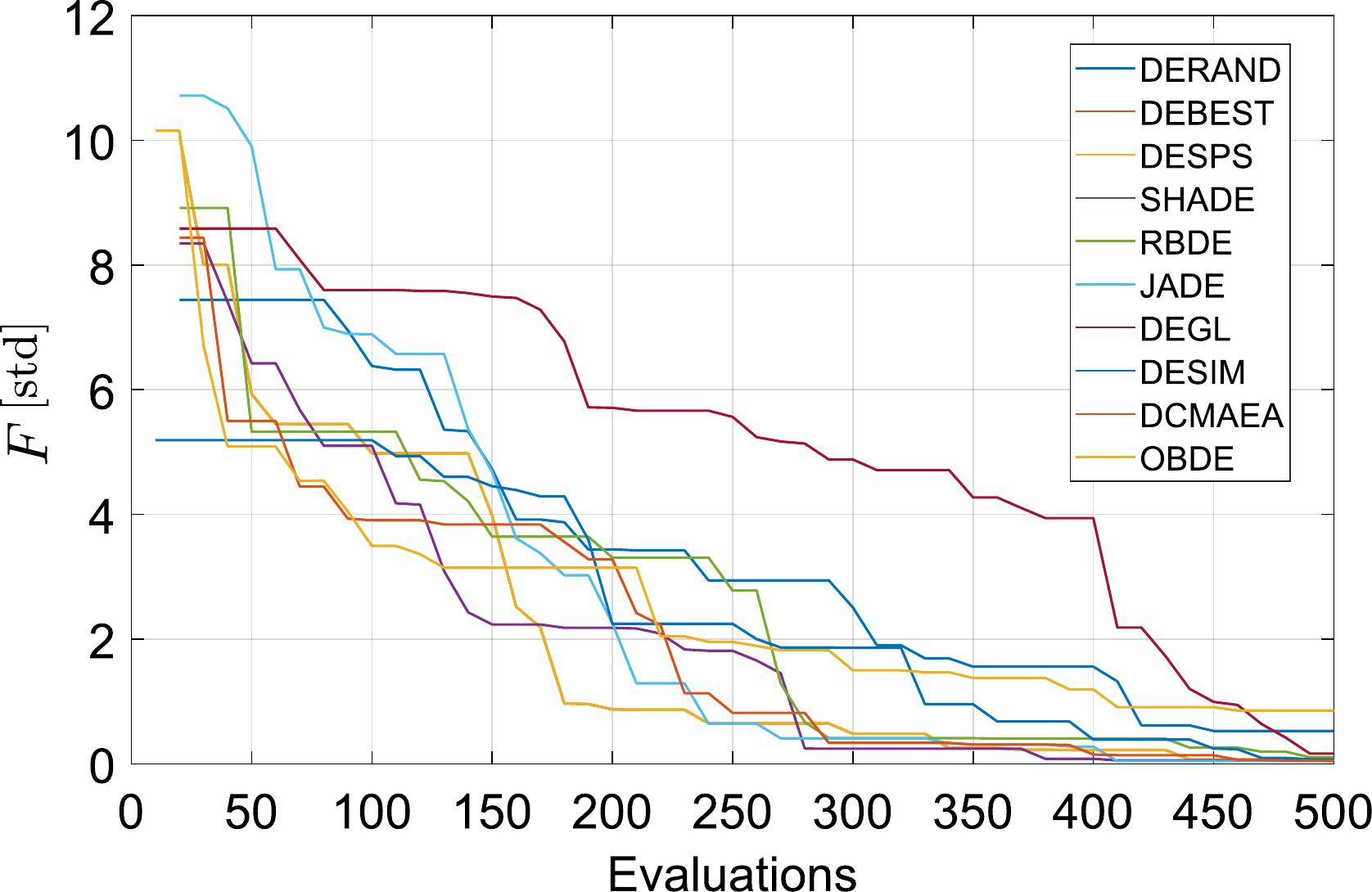}}{Standard Deviation}\hfill
    \stackunder{\includegraphics[width=0.32\textwidth]{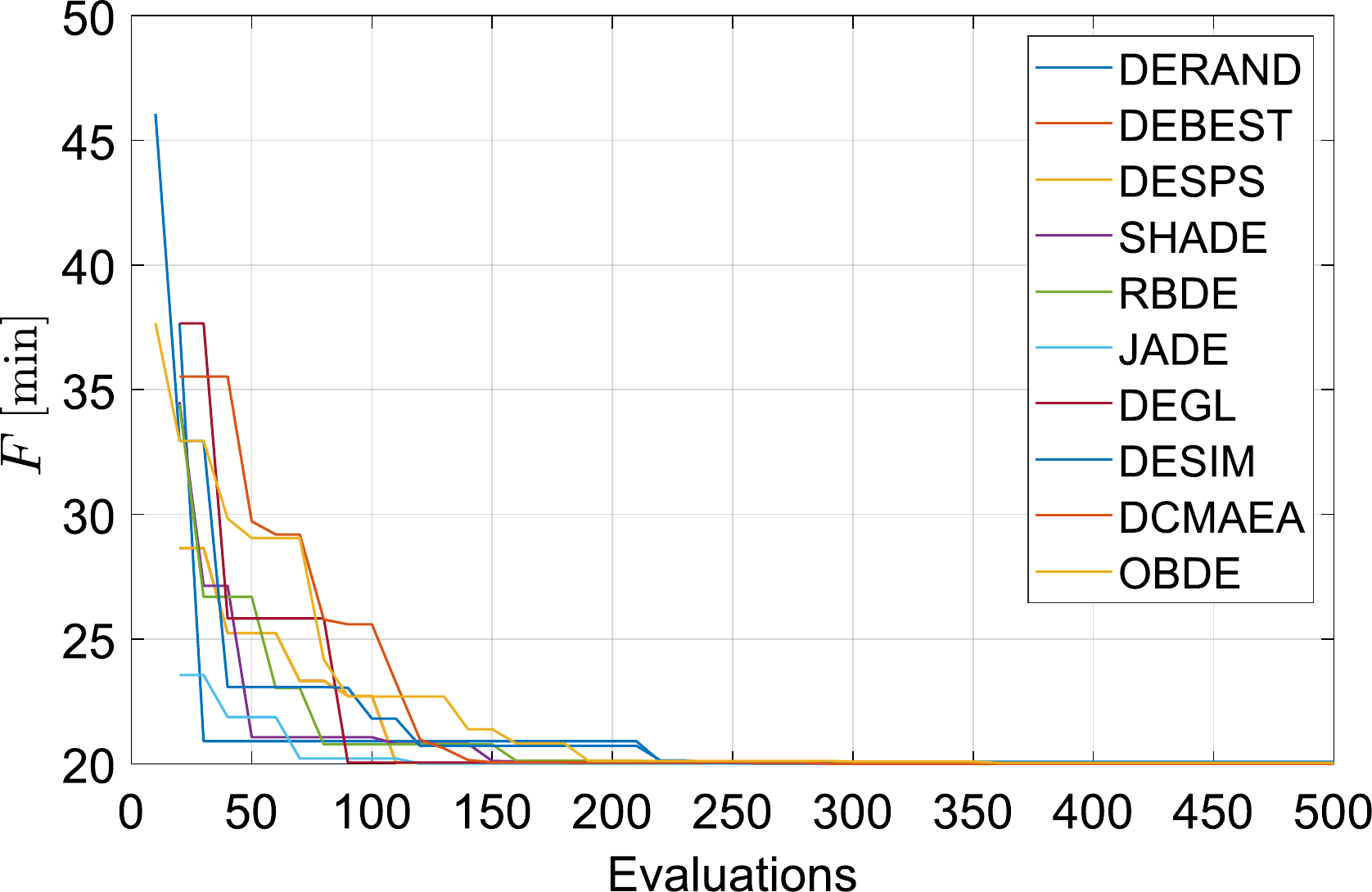}}{Minimum Convergence}\hfill
    \hfill
	\\
    \hline[dotted]
    \textbf{(a)} Scenario 2: $w_1 = 1, ~ w_2 = 1, ~ w_3 = 1$.\\
    \hfill
    \stackunder{\includegraphics[width=0.32\textwidth]{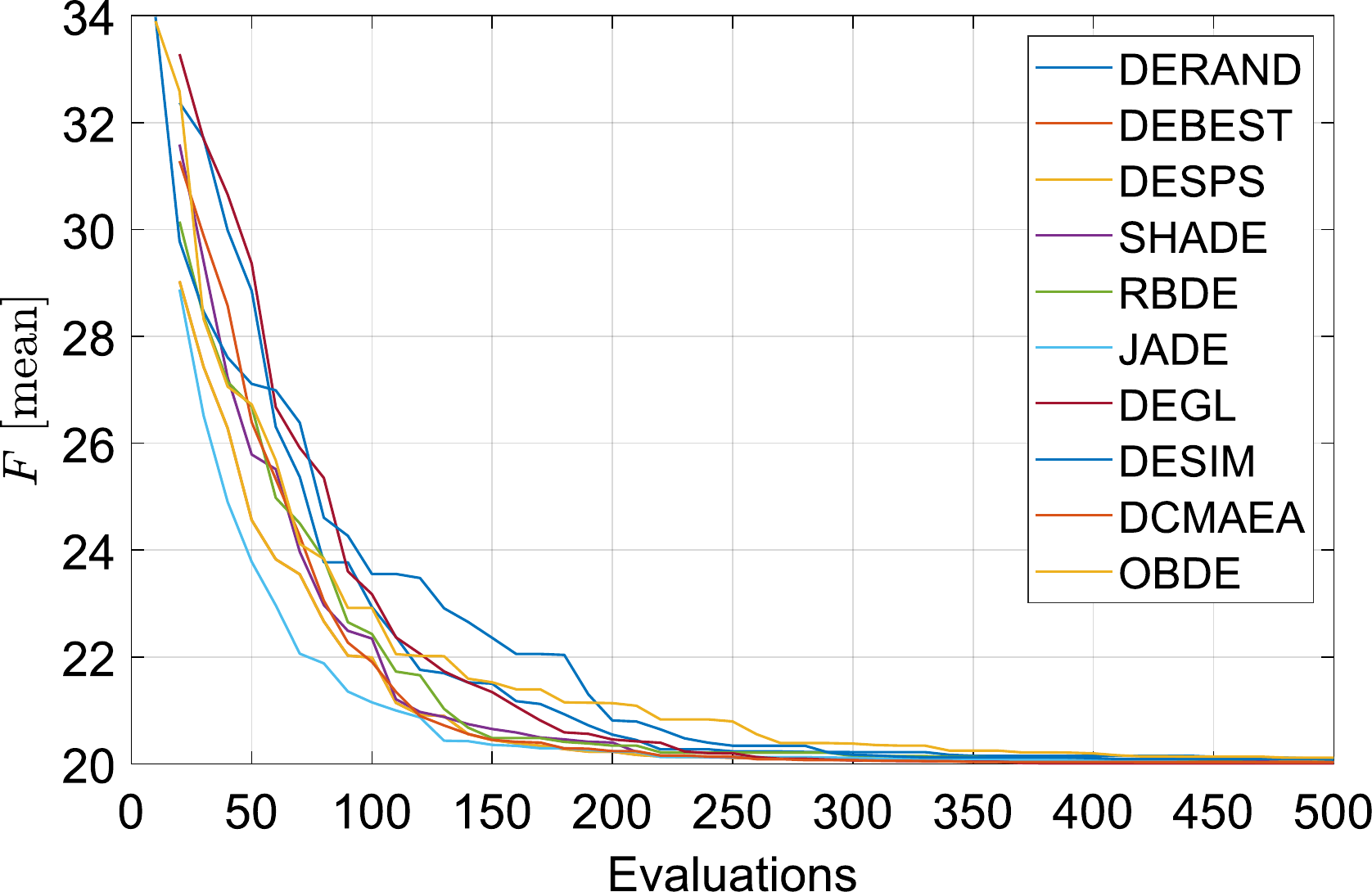}}{Mean Convergence}\hfill
    \stackunder{\includegraphics[width=0.32\textwidth]{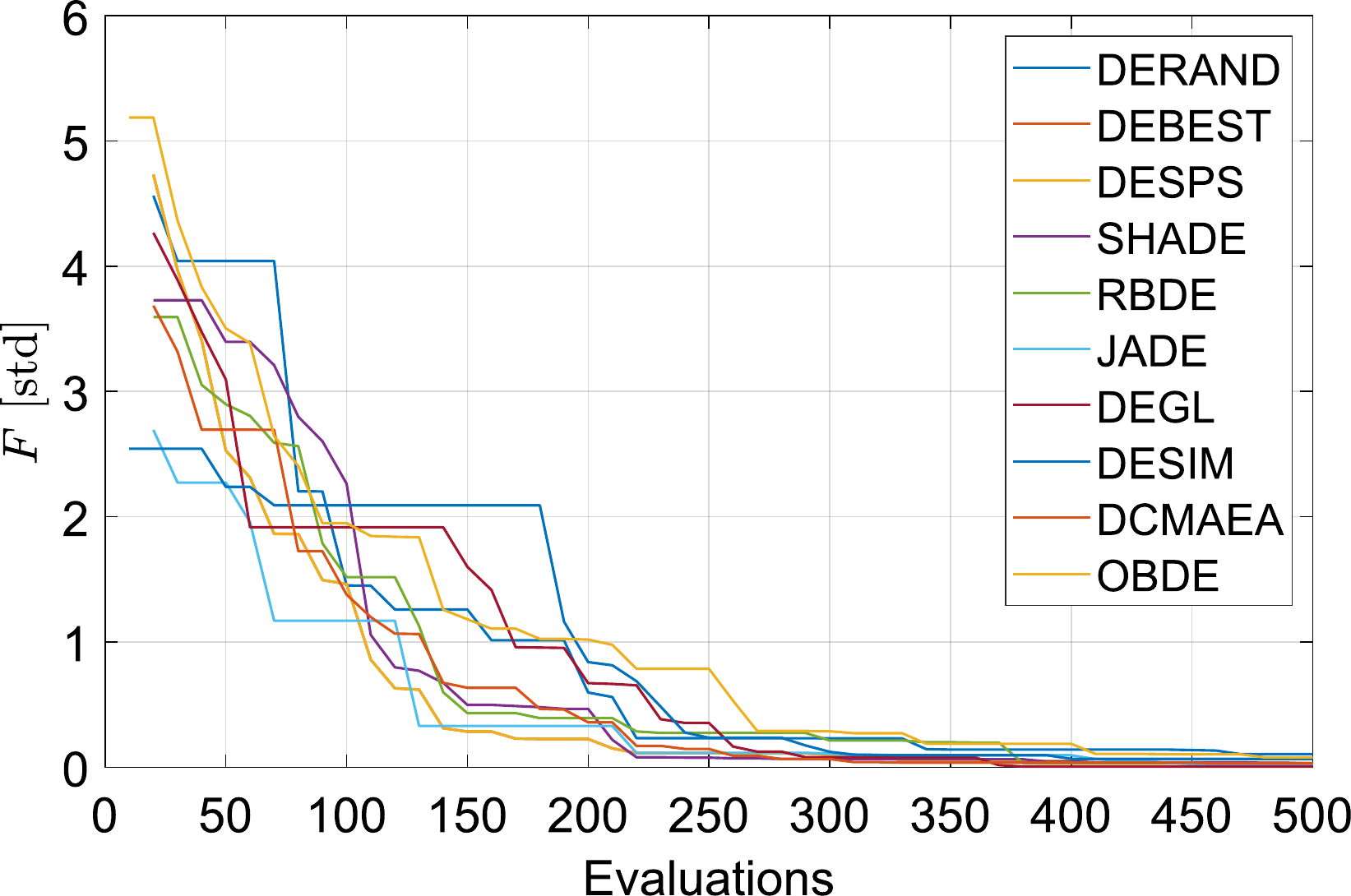}}{Standard Deviation}\hfill
    \stackunder{\includegraphics[width=0.32\textwidth]{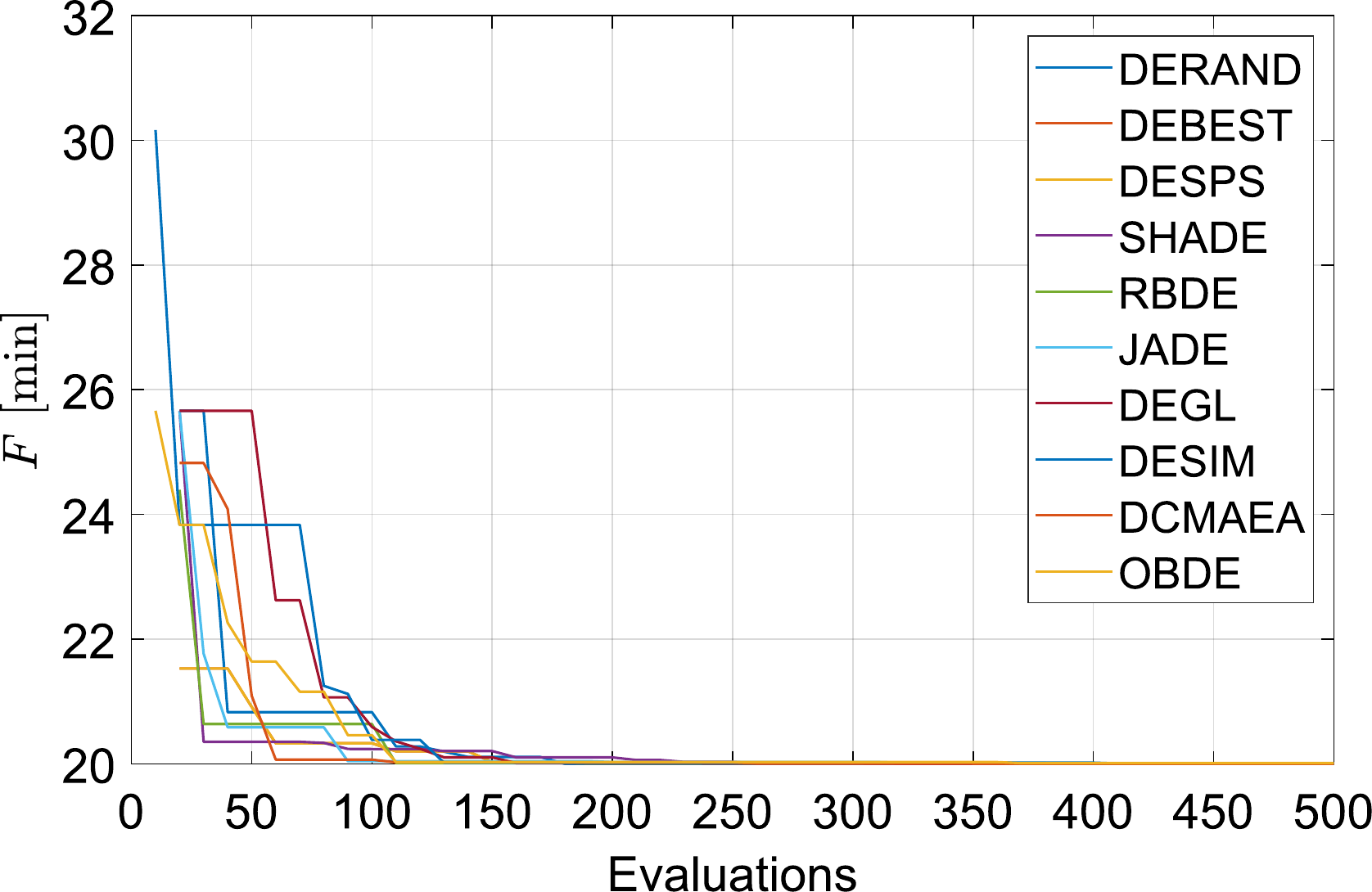}}{Minimum Convergence}\hfill
	\\
    \hline[dotted]
    \textbf{(c)} Overview of the statistical comparisons.\\
    \stackunder{\includegraphics[width=0.48\textwidth]{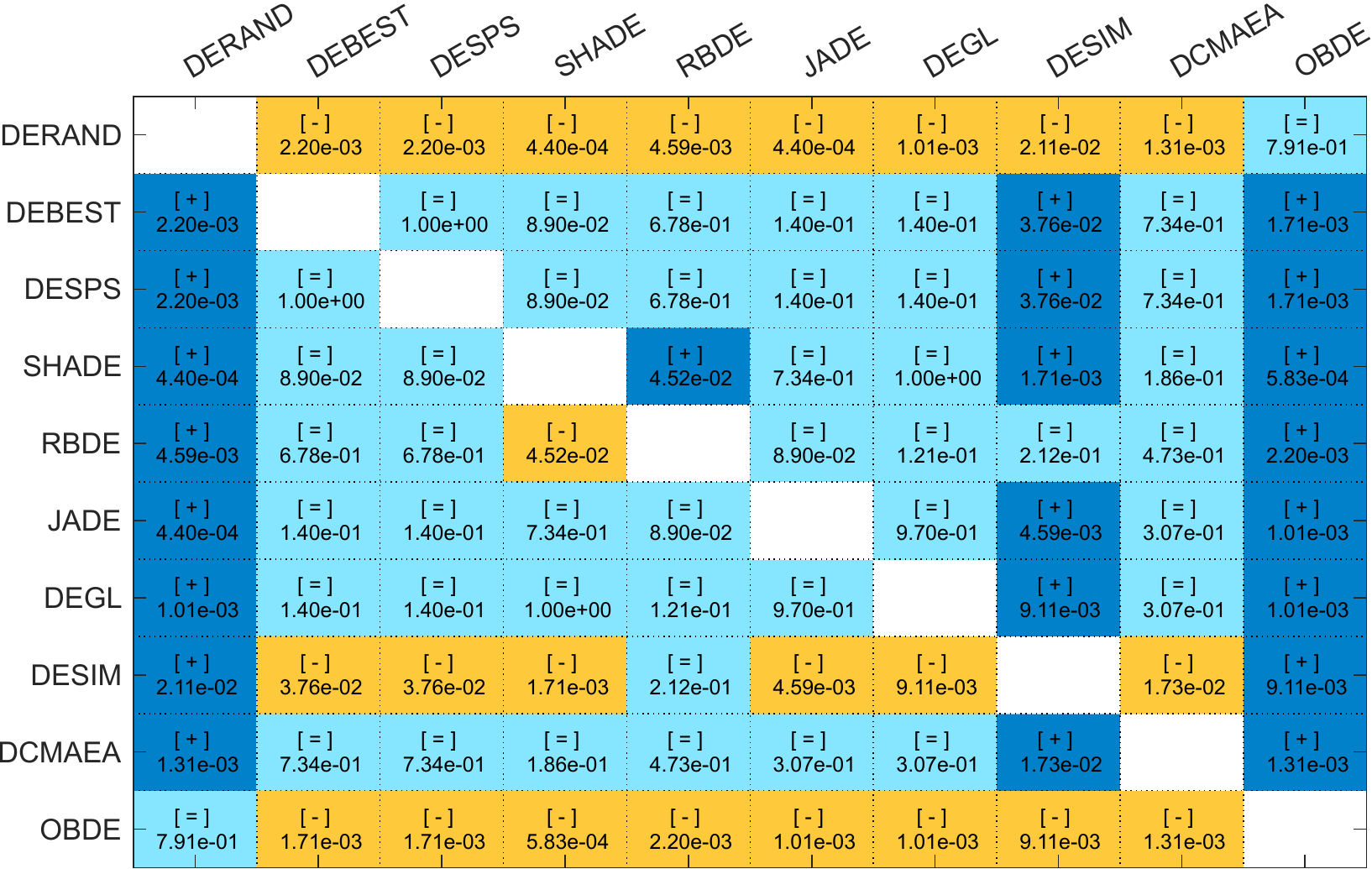}}{Scenario 1}\hfill
    \stackunder{\includegraphics[width=0.48\textwidth]{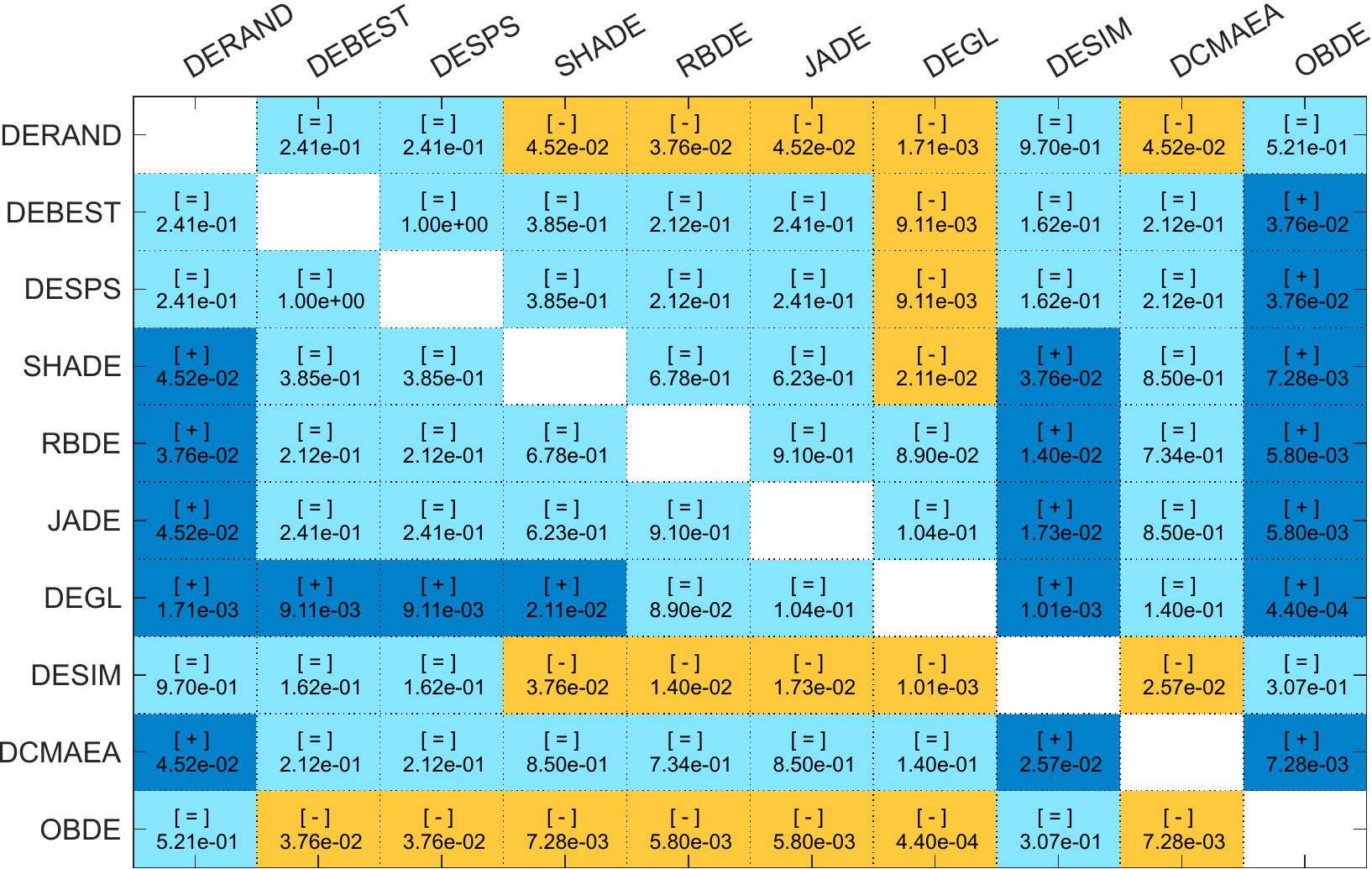}}{Scenario 2}\hfill
    \\
    \hline[0.25em,gray3]
\end{tblr}
\caption{Overview of optimization results: convergences in different scenarios, statistical comparisons. }
\label{opt}
\end{figure*}

Algorithm parameters involve population size 10, probability of crossover $0.5$, scaling factor for difference of vectors $0.7$, and 10 independent runs (due to the stochastic nature of the optimization algorithms). Other parameters followed the suggested values. For a feasibility study, we considered an scenario in which the weight on temperature of the substrate is higher ($w_1 = w_2 = 1, ~ w_3 = 2$), and another scenario for equal weight on inductance and temperature ($w_1 = w_2 = w_3 = 1$). Although mean convergence occurs relatively quick (at 100-300 function evaluations), the standard deviation and the minimum objective functions follow a decreasing trend, implying the feasibility to locate attractive regions of the search space of high-performing power module layouts. Among the studied algorithms, we also note that the class of exploration-focused and diversity-induced algorithms, such as DERAND, OBDE and DESIM underperform their counterparts with exploitation and parameter adaptation schemes (statistical comparisons based on Wilcoxon at 5\% significance level). By looking at the obtained results in Fig. \ref{resnn} - Fig. \ref{opt}, we can observe the feasibility of learning/approximating surrogate functions able to estimate performance metrics related to inductance and temperature of power module layouts, and use those surrogates to optimize/locate feasible and attractive layout configurations of power module substrates. The future work is expected to evaluate and extend the use of surrogates to propose unique/initial configurations for power modules.


%
%
%
%
%

\bibliographystyle{IEEEtran}
\bibliography{mybiblio}

\end{document}